%% file: main.tex
\documentclass{article}

\usepackage{arxiv}

\usepackage[utf8]{inputenc} 
\usepackage[T1]{fontenc}    
\usepackage{hyperref}       
\usepackage{url}            
\usepackage{booktabs}       
\usepackage{nicefrac}       
\usepackage{lipsum}
\usepackage{graphicx}
\usepackage{tikz}
\usepackage{pgfplots}
\usepackage{pgfplotstable}
\usepackage[ruled,linesnumbered]{algorithm2e}
\usepackage{enumitem}
\usetikzlibrary{matrix,chains,positioning,decorations.pathreplacing,arrows,calc,intersections, datavisualization,3d,fit,patterns}
\usepackage{tikz}
\usepackage{import}
\usepackage{mathbbol}
\usepackage{bbold}
\usepackage{amsmath,amsfonts,amssymb}
\usepackage{float}
\usepackage{environ}
\usepackage{pgfplotstable}
\usepackage{setspace}
\usepackage[customcolors]{hf-tikz}
\usepackage{authblk}

\pgfplotsset{compat=newest}

\usepackage[ruled,linesnumbered]{algorithm2e}

\usepackage[colorinlistoftodos]{todonotes}
\usepackage{enumitem}
\subimport{./Include/layers/}{init}

\graphicspath{{./Include/}}

\title{Learning in Convolutional Neural Networks Accelerated by Transfer Entropy}

\begin{document}

\author[1]{Adrian Moldovan}
\author[1]{Angel Ca\c taron}
\author[2]{R\u azvan Andonie}
\affil[1]{Department of Electronics and Computers, Transilvania University, Bra\c{s}ov, 500024, Romania}
\affil[2]{Department of Computer Science, Central Washington University, Ellensburg, WA 98926, USA}
\affil[ ]{adrian.moldovan@gmail.com, cataron@unitbv.ro, razvan.andonie@cwu.edu}

\maketitle

\begin{abstract}
Recently, there is a growing interest in applying Transfer Entropy (TE) in quantifying the effective connectivity between artificial neurons. In a feedforward network, the TE can be used to quantify the relationships between neuron output pairs located in different layers. Our focus is on how to include the TE in the learning mechanisms of a Convolutional Neural Network (CNN) architecture. We introduce a novel training mechanism for CNN architectures which integrates the TE feedback connections. Adding the TE feedback parameter accelerates the training process, as fewer epochs are needed. On the flip side, it adds computational overhead to each epoch.  According to our experiments on CNN classifiers, to achieve a reasonable computational overhead--accuracy trade-off, it is efficient to consider only the inter-neural information transfer of a random subset of the neuron pairs from the last two fully connected layers. The TE acts as a smoothing factor, generating stability and becoming active only periodically, not after processing each input sample. Therefore, we can consider the TE is in our model a slowly changing meta-parameter.
\end{abstract}


\section{Introduction}
Sometimes, it is difficult to distinguish causality from statistical correlation. A prerequisite of causality is the time lag between cause and effect: the cause precedes the effect~\mbox{\cite{Shadish2001, Marwala2015}}. We consider here causality in a statistical sense, measured by  information transfer. Statistical causality direction is inferred from the knowledge of a temporal structure, assuming that the cause has to precede the effect \cite{Zaremba2014}. According to  the authors of \cite{Lizier2010}, causal information flow describes the causal structure of a system, whereas information transfer can then be used to describe the emergent computation on that causal structure. For practical reasons, it is convenient to accept that causality can be measured by information transfer, even if the two concepts, are not exactly the same.

Transfer Entropy (TE) is an information transfer measure introduced by Schreiber~\cite{Schreiber2000} as a measure used to quantify the statistical coherence between events (usually, time series). Later, TE was considered in the framework of Granger’s causality \cite{Barnett2009, Hlavackova-Schindler2011}. Typically, causality is related to whether interventions on a source have an effect on the target, whereas information transfer is related to whether observations of the source can help predict state transitions on the target. According to Lizier et al. \cite{Lizier2010}, to be considered a correct interpretation of information transfer, TE should only be applied to causal information sources for the given destination.  A comparison between the TE and causality indicators can be found, for instance, in  \cite{Lizier2010}. With this in mind, we will use the information transfer (measured by the TE) to establish the presence of and quantify causal relationships.

Massey \cite{Massey1990} defined the directivity of information flow through a channel in the form of directed information. In the presence of feedback, this is a more useful quantity than the traditional mutual information. Similarly, the TE measures the information flow from one process to another by quantifying the deviation from the generalized Markov property as a Kullback--Leibler distance. Therefore, the TE can be used to estimate the directional informational interaction between two random variables. 

In our case, we quantify the information transfer between the neural layers of feedforward neural architectures. The information between these layers is directed: during the feedforward phase of the backpropagation algorithm, the layers are computed successively. We reduce the weights increment for larger values of TE with the objective of preserving the network configuration if the information is efficiently transferred between layers. The cause (the output from a neural layer) precedes the effect (the input to a subsequent layer). We measure causality by this directional information transfer. The directed informational transfer between two neural layers cannot be measured by a symmetrical statistical measure, such as mutual information, where cause and effect simultaneously coexist.

A variation of the TE is the Transfer Information Energy (TIE), introduced by us in~\mbox{\cite{Cataron2017, Cataron2019}} as an alternative to the TE. The TE measures of the reduction in uncertainty about one event given another, whereas the TIE measures the increase in certainty about one event given another. The TE and the TIE can be both used as quantitative indicators of information transfer between time series. The TIE can be computed \textasciitilde20\% faster than the TE \cite{Cataron2017}.

Recently, there is a growing interest in applying TE in quantifying the effective connectivity between artificial neurons \cite{Lizier2011, Vicente2011, Shimono2015, HuiFang2018}. For instance, the reservoir adaptation method in~\cite{Obst2010} optimizes the TE at each individual unit, dependent on properties of the information transfer between input and output of the system. Causal relationships within a neural network were defined in \cite{Feraud2002}. It is a natural question if causal relationships quantified by TE information transfer measure can be inferred from neural networks and included in the learning mechanisms of neural architectures. There are few results reporting applications of TE in training neural networks: Herzog et al. \cite{Herzog2017, Herzog2020}, Patterson et al. \cite{Patterson2017}, \mbox{Obst et al. \cite{Obst2010}}, and  Moldovan et al.  \cite{Moldovan2020}.

Herzog et al. \cite{Herzog2017} computed the feedforward TE between neurons to structure neural feedback connectivity. These feedback connections were used in the training algorithm of a  Convolutional Neural Network (CNN) classifier. Herzog et al. averaged (by layer and class) the calculation of TE gathered from directly or indirectly connected neurons, using thresholded activation values. The averaged TEs were then used in the subsequent neuron's activations, within the training algorithm. Only one TE derived value is used for each of the layers. Herzog et al. observed that there is a decreasing feedforward convergence towards higher layers. Furthermore, the TE value is in general lower between nodes at larger layer distances than between neighbors. This is caused by the fact that long-range TE is calculated by conditioning on the intermediate layers. 

Herzog et al. continued their research in  \cite{Herzog2020}. Their goal was to define clear guidelines about how to compute the TE based neural feedback connectivity to improve the overall classification performance of feedforward neural network classifiers. Structuring such feedback connection in a deep neural model is very challenging because of the very large number of candidate connections. For AlexNet, Herzog et al. narrowed the TE feedback connections to about 3.5\% of all possible feedback connections. Then they used a genetic algorithm to modify their connection strengths and obtain in the end a set of very small weights, similar to many feedback paths in the brain, which amplify already connected feedforward paths only very mildly. According to their experiments in \cite{Herzog2020}, this technique improved the classification performance. In a nutshell, the algorithmic steps in \cite{Herzog2020} are (i) train the network employing standard backpropagation, (ii) fine-tune the resulted network using feedback TE connections, and (iii) apply a genetic algorithm to generate the best performing network.

Inspired by Herzog et al.'s paper \cite{Herzog2017}, we defined in \cite{Moldovan2020} a novel information-theoretical approach for analyzing the information transfer (measured by TE) between the nodes of feedforward neural networks. The approach employed a different learning mechanism than the one in \cite{Herzog2017}. The TE was used to establish the presence of relationships and the quantification of these between neurons and the TE values were obtained from neuron output pairs located in consecutive layers. Calculating the TE values required a series of event values that were obtained by thresholding the neurons' outputs with a constant value. We introduced a backpropagation-type training algorithm which used TE feedback connections to improve its performance. Compared with a standard backpropagation training algorithm, the addition of the TE feedback in the training scheme implies a computational overhead needed to compute the TE values in the training stage. However, according to our experiments, adding the TE feedback parameter has three benefits \cite{Moldovan2020}: (a) it accelerates the training process---in general, less epochs are needed; (b) generally achieve a better test set accuracy; and (c) it generates stability during training. The neural models trained in \cite{Moldovan2020} were relatively small. This allowed to use all training samples to compute the TE from time series. When training complex models with large datasets, for computational reasons, this is unpractical. The question (and main motivation for our current work) is how to adapt our technique to such real-world cases. 

We extend here the results from \cite{Moldovan2020} and adapt them to a much larger neural architecture---the CNN network. Rather then being a simple generalization, it is a novel approach, as we had to redefine the network training process. The motivation is not only the popularity of CNNs in current deep learning applications, but also the fact that despite  the  ability  of  generating  human-alike  predictions,  CNNs  still  lack  a  major  component:  interpretability.  CNNs utilize a hierarchy of neural network layers.  We consider that the statistical aspects of information transfer between these layers can bring an insight into the feature abstraction process. Our idea is to use TE as a momentum factor in the backward step of backpropagation of error and update the weights in accordance with the uni-directional amount of information transferred between pairs of neurons. We thus leverage the significant informational connection between two units in the learning phase, obtaining a better granularity of the weights' increments.

In contrast to the work in \cite{Herzog2020}, we integrate the TE feedback connections in the training algorithm, and do not use them in a subsequent fine-tuning. The way we select the feedback connections is also different. The similarity between Herzog et al.'s method and our work consists in using TE to compute neural feedback connections. However, the two approaches are very different. Using feedback connections, a general training algorithm like backpropagation adds a computational overhead, not discussed in \cite{Herzog2017, Herzog2020}. We may expect that there is a trade-off between execution time and classification accuracy. 

Beyond the exploratory aspect of our work, our main insights are twofold.  First, we attempt to improve (training time, accuracy) the training algorithm of CNN architectures. Second, we create the premises for further information transfer interpretation within deep learning models. 

The rest of the paper is structured as follows. Section \ref{background} introduces the TE definition and notations, whereas  Section \ref{howto} explains how we compute the TE feedback in a CNN. \mbox{Section  \ref{integrate}} introduces the core of our novel method---the integration of the TE feedback mechanism into the training phase of a CNN. Experimental results are presented in \mbox{Section \ref{results}}. \mbox{Section \ref{conclusions}} contains the final remarks and open problems.

\section{Transfer Entropy Notations} \label{background}

The recent literature on TE applications is rich and illustrates an increasing interest for the use of this tool in a broad range of applications to measure the directional information flow between time series based on their probability density functions. Applications of TE to date has mainly been concentrated in neuroscience, bioinformatics, \mbox{artificial life}, climate science, finance, and economics \cite{Bossomaier2016}. An introduction to TE is offered by \cite{Bossomaier2016}. A nonparametric approach of causality considering the conditional entropy was introduced by Baghli \cite{Baghli2006}. An extensive analysis of causality detection based on information-theoretic approaches in time series analysis is provided in \cite{Hlavackova-Schindler2007}. 

 To introduce the formal definition of TE, let us consider two discrete stationary processes $I$ and $J$. Relative to time $t$, $k$ previous data points of process $I$, $l$ previous data points of process $J$, and the next point of process $I$ contribute to the definition of TE as follows~\cite{Schreiber2000, Kaiser2002}:
\begin{equation}\label{eq:TEcond}
	TE_{J\rightarrow I}=\sum_{t=1}^{n-1}{p(i_{t+1},i_{t}^{(k)},j_{t}^{(l)}) \: log \frac{p(i_{t+1}|i_{t}^{(k)},j_{t}^{(l)})}{p(i_{t+1}|i_{t}^{(k)})}}, 
\end{equation}
where $i_t^{(k)}$ and {$j_t^{(l)}$} are the $k$ and $l$ dimensional delay vectors of time series $I$ and $J$, respectively, and $i_{t+1}$ and $j_{t+1}$ are the discrete states at time $t+1$ of $I$ and $J$, respectively. The generalization of the entropy rate to two processes is a method to obtain a mutual information rate which measures the deviation from independence, and therefore $TE_{J\rightarrow I}$ can be obtained from Kullback entropy \cite{Schreiber2000}. TE provides an evaluation of the asymmetric information transfer between two stochastic variables, being a tool which can be used to estimate the unidirectional interaction of pairs of time series. 

The precise calculation of the entropy-based measures is an well-known  difficult task and the computational effort is still significant when accurate estimation of TE from a dataset is required \cite{Gencaga2015}. One of the most widely used approach is based on the histogram estimation with fixed partitioning, bu it is not scalable and is sensible to bins width setting. As TE is derived from the nonparametric entropy estimation, popular methods are widely used for computing the transfer entropy \mbox{TE \cite{Hlavackova-Schindler2009,Gencaga2015,Zhu2015}}: kernel density estimation methods, nearest-neighbor, Parzen window density estimation, etc.

\section{Computing the TE Feedback in a CNN} \label{howto}
CNNs employ a particular form of linear transformation: \emph{convolution}. A convolution operation retains the important and variational features of an input, while flattening the non-variant components. CNN design follows vision processing in living organisms and became very popular starting with the seminal paper of Yann LeCun et al. \cite{726791} on handwritten character recognition, where they introduced the LeNet-5 architecture. Since then, research on CNN architectures produced a variety of deep networks, like AlexNet \cite{Alex2012}, VGG \cite{Simonyan15}, GoogleNet \cite{DBLP:journals/corr/SzegedyLJSRAEVR14}, ResNet \cite{DBLP:journals/corr/HeZRS15}, and more recently EfficientNet \cite{DBLP:journals/corr/abs-1905-11946}. These have many applications which makes them widely used in image recognition and classification, medical imaging, and time series analysis, to name a few. 

Despite the ability of generating, especially in image recognition tasks, human-alike predictions, CNNs still lack a major component: interpretability \cite{Musat2020}. Neural networks in general are known for their black-box type of behavior, hiding the inner working mechanisms of reasoning. However, reasoning and causal explanations are extremely  important for domains like medicine, law, and finance. It is tempting to model statistical cause--effect relationships between CNN neural layers using TE, with the goal to contribute to the interpretability of a CNN. As a first step, even improving the learning algorithm using an information transfer indicator is interesting, as it may lay the ground for future causality explanations.

To quantify inter-neural information transfer in a CNN, we have to quantify the relationship between training the samples and the output values of neurons. The measurable relationship is constructed by selecting subsequent layers and extracting TE values for the pairs of neurons implied. The computed TE values will directly participate in learning mechanism of the network, described in Section \ref{integrate}.

Each TE value is computed by combining two time series, $I$ and $J$, each obtained by binarization of the activation function of a neuron, with threshold $g$. Each value in time series $I$ and $J$ is a neuron output computed for an input sample. An ideal binarization threshold should produce only few positive values. The reason is that, in such a case, the obtained TE values tend to have a comparative value with learning rate,  and then tend to flatten during CNN training; this gives stability to the learning process. A similar binarization technique was used by Herzog et al. in \cite{Herzog2017, Herzog2020}.

We compute  $I$ and $J$ in Equation \eqref{eq:TEcond} for individual pairs of neurons from adjacent layers $k$ and $l$, $l$ being the next layer after $k$. Index $t$ is the position of an input sample in the training sequence. Time series $I$ and $J$ are updated online, after processing each input sample. For each considered pair of neurons, the TE value is computed only periodically, after a processing a fixed number of training samples. For all pairs of neurons in layers $k$ and $l$, we obtain a triangular adjacency matrix of TE values.

It is computationally not feasible to compute the TE values for all possible pairs of neurons. Actually, according to our experiments, not all inter-neural information transfers are relevant for the training process of a network. We observed that the highest impact of using TE in training a CNN is within the last layers. We interpret this as a fine tuning of the classification process, as the classifiable features are available in the final layers of the network. Therefore, we compute the TE values only between the neurons of the last two layers of the CNN. The focus on the last two layers diminishes the computational overhead for calculating the TE values. Our approach is different than the one in \mbox{Herzog et al. \cite{Herzog2017, Herzog2020}}, where TE interactions between non-adjacent layers are also calculated.

\section{TE Feedback Integration in CNN Training} \label{integrate}

Backpropagation  training has two standard phases \cite{Rumelhart1986}: feedforward and backward. In the forward phase, for each input sample, in addition to the activations of the neurons for each layer, we also record the time series needed for the TE computation.

The last layer's output is used to calculate the error, which for classification tasks is usually $L=-ln(p_c)$. In the backward phase the weights of error are updated, in reverse order, starting from the last layer. In contrast to the standard backpropagation algorithm, we update the weights with a value resulted from multiplying the current weights by the identity matrix minus the computed TE values. The two phases (feedforward and backward) are alternated until the entire training set is used, and this completes one training epoch. The training consists of several epochs, the  training set being randomized at the start of each epoch. In practice, the backpropagation algorithm is used in conjunction with a Mini-batch  Gradient Descent (SGD) \cite{Shalev2011} that updates the weights after a batch of training samples is processed. Mini-batch Gradient Descent is the algorithm of choice for neural networks training. Updating the gradient with the TE addition is synchronized with the batch gradient updates. Without synchronization, the SGD method will diminish the impact of the TE term.

After each batch of training samples goes through forward computation, the backpropagation of errors and new weights computation is performed using Algorithm \ref{alg:tebackprop}. Different than in the standard backpropagation algorithm \cite{Rumelhart1986}, we multiply (line 8 of the algorithm) the updated weights by $(\bf{I} - (\mathbf{te}^{(k)})^\top)$.

Each pair of neurons generates a \emph{te} value; two consequent layers \emph{k} and \emph{l} will produce triangular matrix  $\mathbf{te}^{(k)}$, used to update the weights for layer \emph{k} as shown in the \mbox{Algorithm \ref{alg:tebackprop}}. From line 8 of the algorithm, we conclude that the right member will tamper the weights values, in particular when the cost function has a strong coercive action on the misclassified samples. This accelerates the learning process, as the weights are updated with smaller deltas at the beginning of epochs. Without \emph{te}, the weights receive larger corrections at the beginning of the epoch.

\begin{algorithm}
\setstretch{1.35}
\LinesNumbered
\DontPrintSemicolon
\SetKwData{Epoch}{e}
\SetKwData{Batch}{b}
\SetKwData{MaxEpochs}{R}
\SetKwData{Layer}{k}
\SetKwData{Sample}{sample}
\Begin{
    \ForEach{layer k=$1, 2, ..., l$}{
        $\pmb{z}^{(k)}=\pmb{w}^{(k)}\pmb{a}^{(k-1)}+\pmb{b}^{(k)}$, where $\pmb{z}^{(k)}$ is the output of layer $k$ and $\pmb{b}$ is the bias\;
        $\pmb{a}^{(k)}=\sigma(\pmb{z}^{(k)})$, where $\pmb{a}^{(k)}$ is the output of the activation function $\sigma$\;
    }
    \ForEach{layer k=$l-1, ..., 1$}{
        $\delta^{(k)}=((\pmb{w}^{(k+1)})^\top\delta^{(k+1))}\odot\sigma'(\pmb{z}^{(k)})$, $\delta^{(k)}$ is the layer error, $\odot$ is the elementwise product\;
        $\pmb{w}^{(k)}=(\pmb{w}^{(k)}-\eta\delta^{(k)}\pmb{a}^{(k-1)})\cdot(\mathbf{I} - (\pmb{te}^{(k)})^\top)$ where $\mathbf{I}$ is the identity matrix\;
    }
}
\caption{Backpropagation using TE for a single step and a single mini-batch. Mini-batches are obtained by equally dividing the training set by a fixed number. This algorithm is repeated for all the available mini-batches and for a number of epochs. Bold items denote matrices and vectors. $\sigma'$ is the derivative of the activation function $\sigma$. The $k$ and $l$ indices are the same as the ones in \mbox{Equation \eqref{eq:TEcond}}}
\label{alg:tebackprop}
\end{algorithm}

\clearpage
\section{Experimental Results} \label{results}

All experiments were completed on an AWS Sagemaker ml.p2.xlarge instance using NVIDIA Tesla K80 GPU, with 4 vCPU cores and 61 GB of RAM, engineered on top of PyTorch 1.7.1. The repository is available on GitHub 
 \hyperlink{https://github.com/avmoldovan/CNN-TE}{https://github.com/avmoldovan/CNN-TE}. 

We used the following well-known datasets as benchmarks: CIFAR-10 \cite{CIFAR10Kriz2009}, FashionMNIST \cite{fashionMNIST2017}, STL-10 \cite{STL102011}, SVHN \cite{SVHN2011}, and USPS \cite{uspsdataset} datasets. This selection was determined by the vast amount of literature surrounding it and the number of available implementations and comparisons.

The networks used consist of the following sequential components: convolution and feature engineering, deconvolution and classifier (in this order). Within these, various mechanisms were used to prevent overfitting (e.g., dropout) and obtain normalization.

Our experiments and additions to this architecture involved mainly the classifier part of the network, but we have also ran experiments on the convolutional layers.

\begin{figure}
\includegraphics[width=12.0cm]{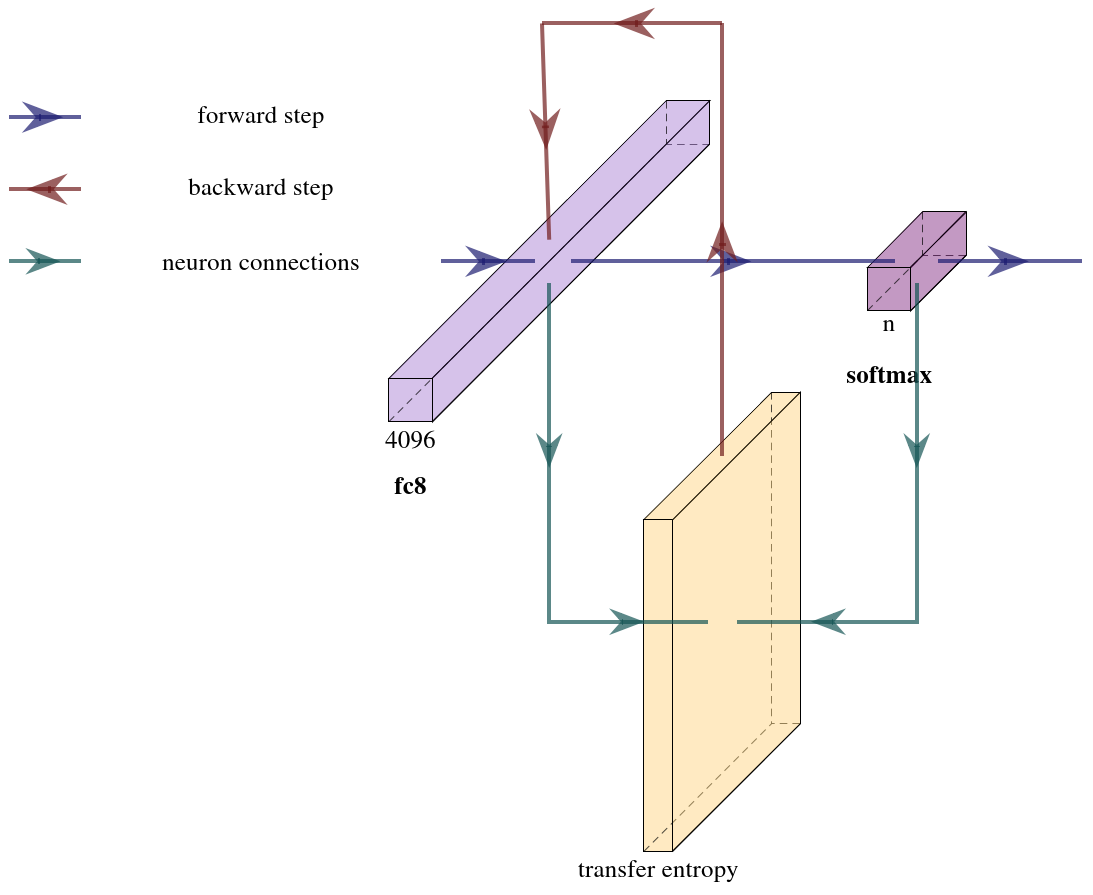}
\label{fig:teapply}
\caption{During the feedforward step, we compute time series $I$ and $J$, and the $\bf{te}$ matrix, as shown by the green arrows. When the backward step propagates the errors, we then use the $\bf{te}$ matrix in the weight updates as shown in the Algorithm \ref{alg:tebackprop}.}
\end{figure}

We applied the TE on the last (fully connected) two layers---the pre-softmax and softmax layers---with different binarization thresholds determined experimentally (see \mbox{Table \ref{tab:withparams}}). The TE term is applied on the weights of the $k$-th layer (see the red arrows in Figure \ref{fig:teapply}).

The softmax layer transforms the  outputs of a linear layer into probabilities. The maximum probability corresponds to the predicted class. For all outcomes with probability above the threshold, the $J$ time-series are positive. 

\begin{figure}[H]
\includegraphics[width=12.0cm]{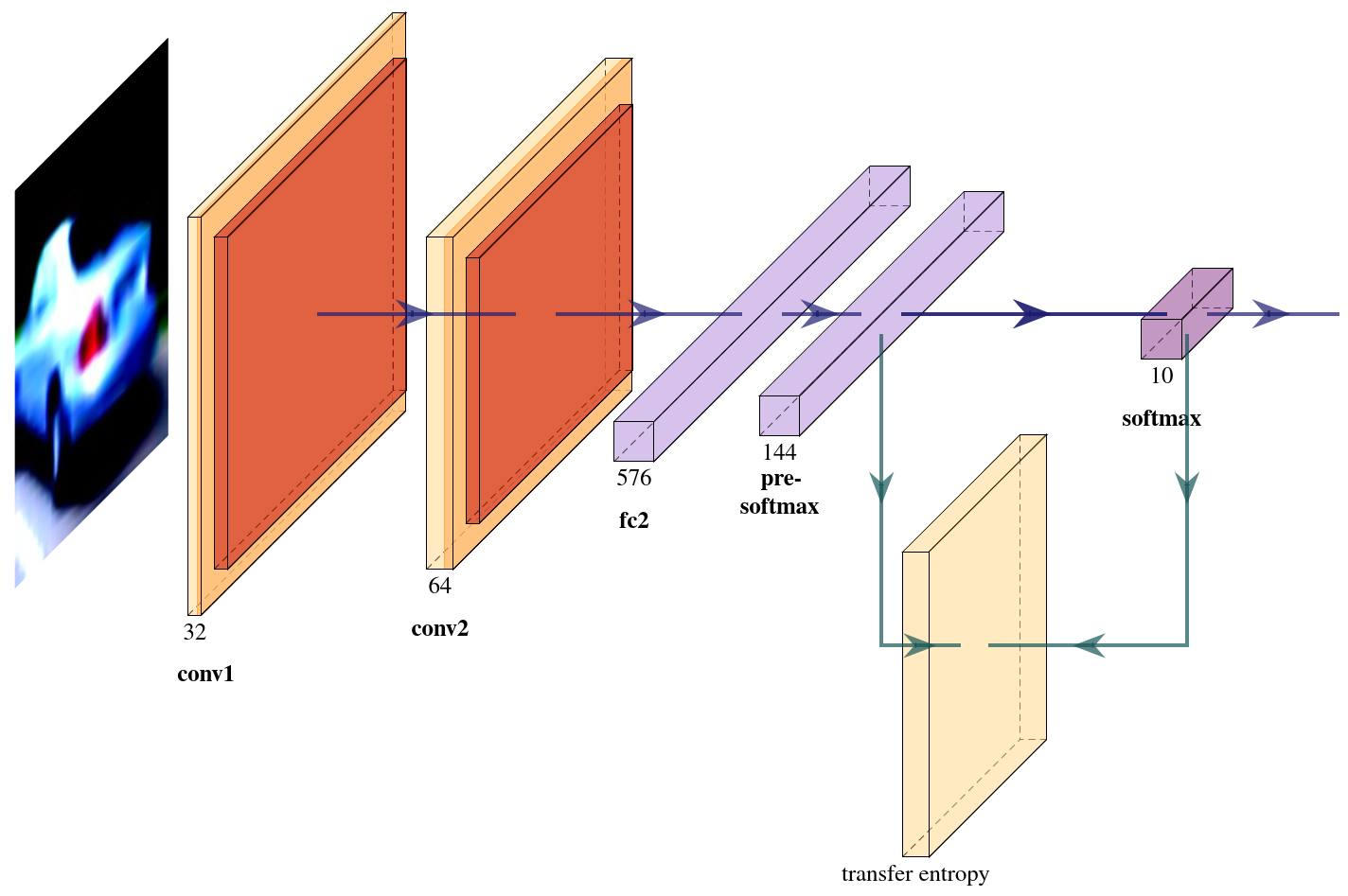}
\caption{Illustration of the feedforward phase for the USPS dataset. The green arrows indicate the layers outputs that are used to compute the TE (Plotted using \url{https://github.com/HarisIqbal88/PlotNeuralNet}).}
\label{fig:uspsarch}
\end{figure}

Figure \ref{fig:uspsarch} depicts the architecture of the network used for the USPS dataset.

\begin{figure}[H]
\includegraphics[width=12.0cm]{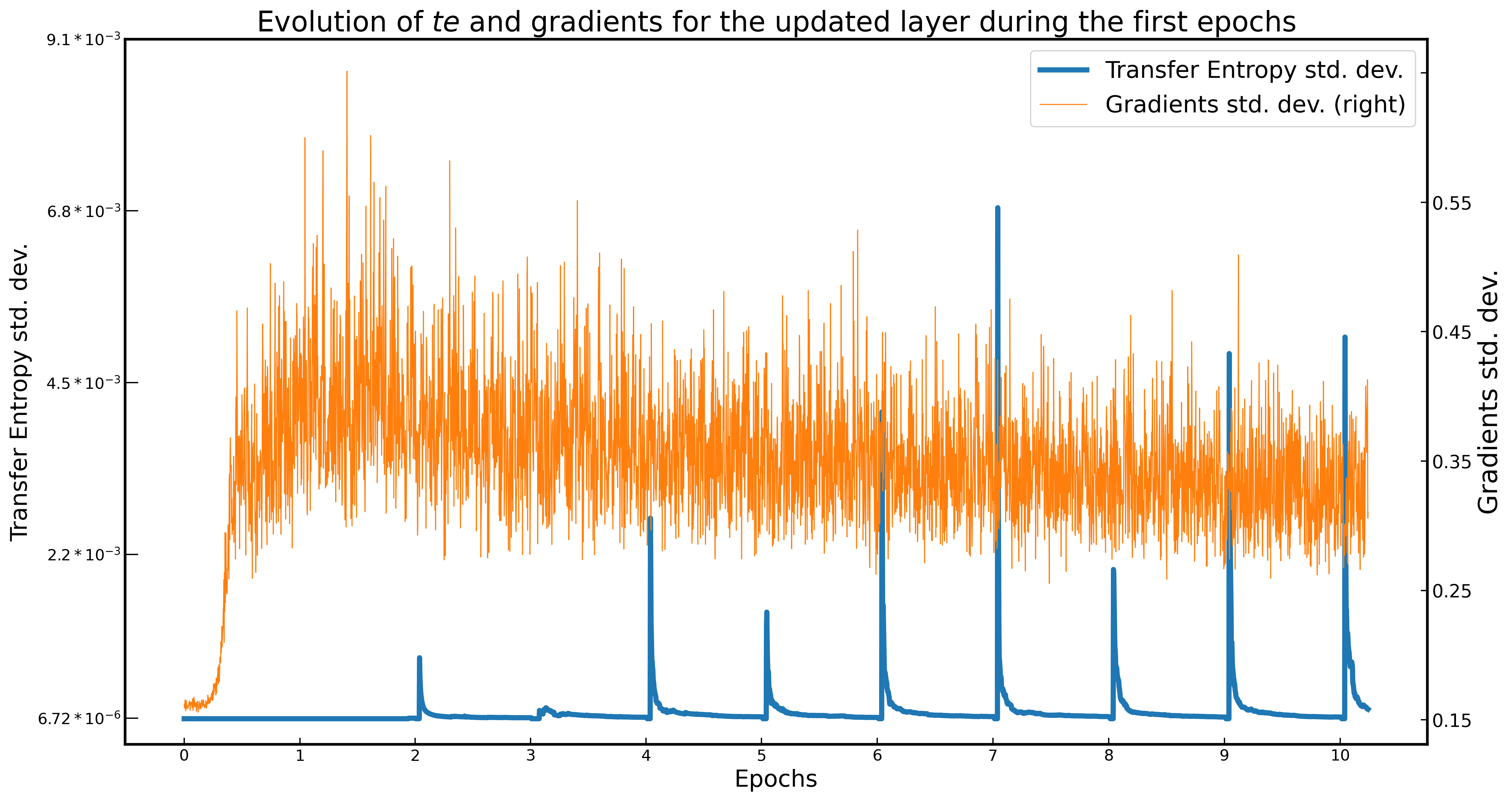}
\caption{Evolution of the $\bf{te}$ standard deviation values on the first 4 epochs for the SVHN+TE dataset, for the pre-softmax layer. Each data point in the plot  represents a batch. The rest of the TE values have a similar shape and decrease slowly during training. We observe the spikes of the TE values at the beginning of each epoch due to the training set randomization. During the first epoch the TE values are not calculated for the first batches in order to prevent anomalous values, thus its value is close to 0.}
\label{fig:tegrads}
\end{figure}

During training, at the beginning of each epoch, we noticed an increased instability, visible through the high variation and values of the gradients, as seen in Figure \ref{fig:tegrads}. These observation apply for all datasets and networks, with or without the TE added. The TE values also exhibit instability and have larger values at the beginning of each epoch. However, the TE values show smaller values during the first epochs due to the selected threshold value that matches larger weights values from subsequent epochs. During each epoch, and also during the whole training process, the slope of the gradients gradually decreases and the TE variation also decreases.

To validate the TE impact, we set a target accuracy to be reached by both implementations with/without TE. We observed the implementation that reaches the target accuracy w.r.t. the number of epochs needed, as well as the average time per epoch. These results show which of the two implementation requires less epochs to reach a target accuracy on the test set. For a fair comparison, we used the same hyperparameters for the CNNs with/without TE implementations (see Table \ref{tab:withparams}). The results are summarized in Tables \ref{tab:cifar10res}--\ref{tab:uspsres}.

\nointerlineskip
\begin{table}
  \caption{Parameters.}
  \tabcolsep=0.4cm
  \begin{tabular}{lrrrrr}
  \toprule
   & \textbf{CIFAR-10+TE} & \textbf{FashionMNIST+TE} & \textbf{STL-10+TE} & \textbf{SVHN+TE} & \textbf{USPS+TE}\\
  \midrule
  learning rate ($\eta$)  & 0.01   &   0.01 &   0.01 &   0.01 &   0.01 \\
  momentum  & 0.9  & 0.9 & 0.9 & 0.9 & 0.9 \\
  dropout     &   0    &   0.25 &     0.  & 0.3   &   0.25\\
  threshold rate 1 ($g_1$)  &   2.0 &   2.0  &   2.0  &   2.0  &   5.0 \\
  threshold rate 2 ($g_2$)  &   0.99 &   0.99  &  0.99  &   0.99  &   0.99 \\
  $te$ window length & 100 & 100 & 4000 & 200 & 90 \\
  batch size    & 500   &   100 &   200 &   200 &   60 \\
  \bottomrule
  \end{tabular}
  \label{tab:withparams}
\end{table}

\begin{table}
  
  \caption{Results for CIFAR-10 \cite{CIFAR10Kriz2009}, with/without TE. The increased training time results from the large size of the last linear layer.}
  \tabcolsep=0.89cm
  \begin{tabular}{lrr}
  \toprule
  \textbf{\space} & \textbf{CIFAR-10+TE}	& \textbf{CIFAR-10}\\
  \midrule
  Target \textbf{\bf{98\%}
} accuracy in epoch    & \textbf{5}		& 6 \\
  Top 1 accuracy at epoch 5  & \textbf{98.02\%}	& 97.58\% \\
  Average epoch duration    &   2110 s    &  \textbf{81 s} \\
  Total training duration   &   10,550 s    &   \textbf{492 s}\\
  \bottomrule
  \end{tabular}
  \label{tab:cifar10res}
\end{table}

\begin{table}
  
  \caption{Results for Fashion-MNIST \cite{fashionMNIST2017} dataset, with/without TE.}
  \tabcolsep=0.59cm
  \begin{tabular}{lrr}
  \toprule
  \textbf{\space} & \textbf{FashionMNIST+TE}	& \textbf{FashionMNIST}\\
  \midrule
  Target \textbf{\bf{97\%}} accuracy in epoch    & \textbf{23}		& 28 \\
  Top 1 accuracy at epoch 23  & \textbf{97.0\%}	& 97.02\% \\
  Average epoch duration    & 71 s   & \textbf{41 s} \\
  Total training duration   &   1720 s    &   \textbf{1162 s}\\
  \bottomrule
  \end{tabular}
  \label{tab:fashionmnistres}
\end{table}
\begin{table}
  
  \caption{Results for STL-10 \cite{STL102011} dataset, with/without TE.}
  \tabcolsep=1.04cm
  \begin{tabular}{lrr}
  \toprule
  \textbf{\space} & \textbf{STL-10+TE}	& \textbf{STL-10}\\
  \midrule
  Target \textbf{\bf{98\%}} accuracy in epoch    & \textbf{5}		& 7 \\
  Top 1 accuracy at epoch 5  & \textbf{98.33\%}	& 78.63\% \\
  Average epoch duration & 28 s & \textbf{7 s} \\
  Total training duration   &   128 s    &   \textbf{53 s}\\
  \bottomrule
  \end{tabular}
  \label{tab:stl10res}
\end{table}
\begin{table}
  
  \caption{Results for SVHN \cite{SVHN2011} dataset, with/without TE.}
    \tabcolsep=1.055cm
  \begin{tabular}{lrr}
  \toprule
  \textbf{\space} & \textbf{SVHN+TE}	& \textbf{SVHN}\\
  \midrule
  Target \textbf{\bf{94\%}} accuracy in epoch    & \textbf{9}		& 11 \\
  Top 1 accuracy at epoch 9  & \textbf{94.05\%}	& 91.67\% \\
  Average epoch duration & 512 s & \textbf{491} s \\
  Total training duration   &   \textbf{4587} s    &   5369 s\\
  \bottomrule
  \end{tabular}
  \label{tab:svhnres}
\end{table}

\begin{table}
  
  \caption{Results for USPS \cite{uspsdataset} dataset, with/without TE.}
  \tabcolsep=1.08cm
  \begin{tabular}{lrr}
  \toprule
  \textbf{\space} & \textbf{USPS+TE}	& \textbf{USPS}\\
  \midrule
  Target \textbf{\bf{99\%}} accuracy in epoch    & 3		& 3 \\
  Top 1 accuracy at epoch 3  & \textbf{99.32}\%	& 99.05\% \\
  Average epoch duration & 376 s   & \textbf{33 s}\\
  Total training duration & 1138 s   & \textbf{102} s\\
  \bottomrule
  \end{tabular}
  \label{tab:uspsres}
\end{table}

\begin{figure}[H]
\includegraphics[width=12.0cm]{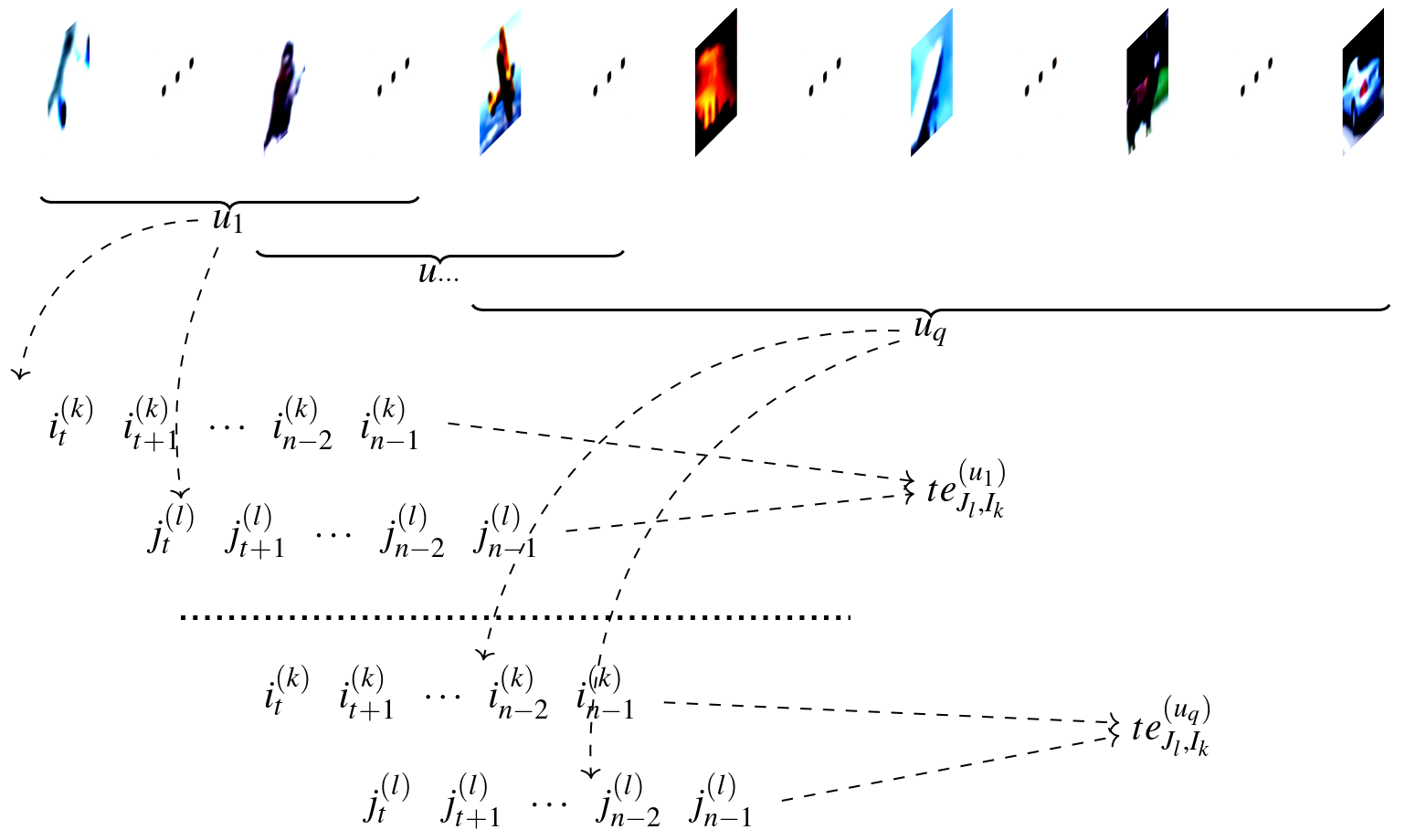}
\caption{Illustration of how time series $I$ and $J$ are produced for a pair of neurons from layers $k$ and $l$, for multiple windows of events $u_1$ \dots $u_q$.}
\label{fig:windowstote}
\end{figure}

As observed in \cite{Moldovan2020}, using time series constructed from the full length of the epoch results in smoother TE values. Computing the TE for large time series (e.g., $>$$10^6$) is computationally impractical. We also observed the necessary length of the time series in that produces observable and positive outcomes. Therefore, we limited this length to $u$ (determined experimentally), using a sliding window technique. To obtain smoother TE values, we slid  the window with every batch of training samples, instead of dividing the training set into $u$-sized windows. Figure \ref{fig:windowstote} illustrates how the time series were computed.

Using this approach, we analyzed the impact of the window length on the accuracy of the classifier. After $q$ training samples, we computed the TE. The $u$ windows overlap partially, as shown in Figure \ref{fig:windowstote}. According to our experiments, limiting the length of the time series does not have a significant impact on the performance of the trained classifier. Furthermore, as we found that a $u$ value five times the batch size is a good trade-off between accuracy and computational overhead.

Convolutional layers are a major building blocks used in CNNs \cite{Lecun99}. A convolution is performed on the input data with the use of a kernel to produce a feature map. Applying the TE to measure the  inter-neural information transfers between the input data and the resulted feature maps is interesting to be considered. We can compute the median of the activations of the neurons within  each convolutional kernel from layer \emph{conv1} (see \mbox{Figure \ref{fig:uspsarch}}) and pair it with the outputs of subsequent layer \emph{conv2}. The obtained $te$  values can be used in the CNN learning process.  In our experiments, under this setup, the learning process diverged. In the best run, the top 1 accuracy hardly reached a considerable value. In addition, this approach has a considerable computational overhead, especially if  we consider several convolutional layers. This justifies our focus on the last two fully connected layers only.

It is also interesting to evaluate the impact the length $s$ of the time series. Experimentally, we observed that when $s$ is a multiple of the batch size $b$, the accuracy maintains a favorable trend. In this scenario, the time series are constructed as illustrated in Figure \ref{fig:windowstote}. The best results were obtained when the length of the series are extended to the full epoch.

In another set of experiments, we tried to minimize the number of considered neuron pairs from the last two layers, with a minimum impact on the achieved accuracy. In other words, we tried to obtain an optimal performance--computational overhead trade-off. We found that the accuracy improves significantly even when using only 10\%  of the randomly selected neurons. The neurons were selected randomly for an entire epoch or by or for a TE window. The two strategies yielded similar results. This is somehow similar to dropout, as only some of the connections are updated using the TE feedback. The top performance is achieved when all neurons are selected.

According to our experiments, using the TE feedback loops for additional layer pairs improves the performance. However, this increases exponentially the number of TE computations needed. For the USPS network (see Appendix \ref{uspslisting}), a very simple dataset, computing the TE for the last two linear layers adds an overhead of ~7 minutes of training time per epoch. Computing the TE only for the pre-softmax and softmax layers training takes ~6 minutes per epoch. Computing the TE for the convolutional layers for the USPS network implies an  increased computational overhead. For all possible pairs of kernels between the convolutional layers, we measured an extra three days of training per epoch. 
Performance cost increases almost linearly  with the number of performed TE calculations. The exact number of TE computations for a pair of layers is a product of the layer sizes and the number of batches. Therefore, it is computationally not practical to compute the TE for all layers.

We also conducted other experiments, applying the TE correction on an identical network, pre-trained without the TE mechanism. We continued training, freezing all layers (except the pre-softmax and softmax layers), and applying the TE correction. The results were not consistent through multiple executions, since the TE training on top of the non-TE training changes the convergence logic and creates instability.

\section{Conclusions and Open Problems} \label{conclusions}

TE can be used to measure how significantly neurons interact \cite{Lizier2011, Vicente2011}. In our study, we add specific TE feedback connections between neurons to improve performance. Our results confirm what we obtained in our previous study on a simple feedforward neural architecture \cite{Moldovan2020}. Adding the TE feedback parameter accelerates the training process, as fewer epochs are needed. On the flip side, it adds computational overhead to each epoch. The optimal balance between these two conflicting factors is application dependent. 

According to our results, in a CNN classifier it is efficient to consider only the inter-neural information transfer of a random subset of the neuron pairs from the last two fully connected layers. The information transfer within these layers has the most significant impact on the learning process, since they are the closest to the high-level classification decision. Many of the inter-neural information transfer connections appear to be redundant, and this allows us to use only a fraction of them. These observations are very interesting and may be further discussed in from a neuroscientific perspective (e.g.,  the vertebrate brain \cite{Gilbert2013, Spillmann2015}).

Generally, to optimize the generalization of a learning algorithm, we try to minimize the number of its parameters. As adding the TE in the learning mechanism generates new hyper-parameters, connected to the integration of the TE in the learning algorithm, the question is if this does not conduct to overfitting and a weaker generalization performance. In our experiments, the TE acted as a smoothing factor, becoming active only periodically, not after processing each input sample. Therefore, we can consider the TE is in our model a slowly changing meta-parameter. This can be related to the hierarchy of quickly-changing vs. slowly-changing parameters in learning neural causal models \cite{Ke2019}. We observed that the TE feedback generates stability during training, this being compliant with the results presented in \cite{Moldovan2020}. 

According to the authors of \cite{Herzog2020}, it is tempting to speculate that a similar principle---an evaluation of the relevance of the different feedforward pathways---might have been a phylo- or ontogenetic driving force for the design of different feedback structures in real neural systems.

A.M., A.C. and R.A. equally contributed to the published work. All authors have read and agreed to the published version of the manuscript.

The costs to publish in open access were covered by Siemens SRL.

This work was published in Entropy journal, Volume 23, Year 2021, Number 9, Article Number 1218, PubMedID 34573843, ISSN 1099-4300,  DOI: 10.3390/e23091218 available also here \hyperlink{https://www.mdpi.com/1099-4300/23/9/1218}{https://www.mdpi.com/1099-4300/23/9/1218}.

The authors declare no conflicts of interest.

\section{Abbreviations}{
The following abbreviations are used in this manuscript:\\

\noindent
\begin{tabular}{@{}p{1 cm} p{12 cm}}
$te$ & transfer entropy value \\
TE & Transfer Entropy \\
$g$ & the binarization threshold \\
$u$ & window of time series used to calculate TE \\
$\mathbf{W}$
 & weights matrix \\ 
$C$ & loss function \\
CNN & Convolutional Neural Network \\
SGD & Stochastic Gradient Descent \\
\emph{CNN+TE} & CNN + Transfer Entropy---our proposed method \\
MLP & Multi-layer perceptron \\
\end{tabular}}




\appendix

\section{CNN Architectures}

\subsection{FashionMNIST}
The architecture used for the FashionMNIST \cite{fashionMNIST2017} dataset is\hyperlink{https://www.kaggle.com/pankajj/fashion-mnist-with-pytorch-93-accuracy}{https://www.kaggle.com/pankajj/fashion-mnist-with-pytorch-93-accuracy}:

\begingroup
\fontsize{8pt}{8pt}\selectfont
\begin{verbatim}
Conv2d(1, 32, kernel_size=(3, 3), stride=(1, 1), padding=(1, 1))
BatchNorm2d(32, eps=1e-05, momentum=0.1, affine=True, track_running_stats=True)
ReLU(inplace=True)
MaxPool2d(kernel_size=2, stride=2, padding=0, dilation=1)
Conv2d(32, 64, kernel_size=(3, 3), stride=(1, 1))
BatchNorm2d(64, eps=1e-05, momentum=0.1, affine=True, track_running_stats=True)
ReLU(inplace=True)
MaxPool2d(kernel_size=2, stride=2, padding=0, dilation=1)
Linear(in_features=2304, out_features=600, bias=True)
Dropout(p=0.25, inplace=False)
Linear(in_features=600, out_features=120, bias=True)
Linear(in_features=120, out_features=10, bias=True)
Softmax(dim=1)
\end{verbatim}
\endgroup

\subsection{CIFAR-10}\label{cifar10listing}
For CIFAR-10 \cite{CIFAR10Kriz2009} dataset we used the following layout \hyperlink{https://github.com/aaron-xichen/pytorch-playground}{https://github.com/aaron-xichen/pytorch-playground}:
\begingroup
\fontsize{8pt}{8pt}\selectfont
\begin{verbatim}
Conv2d(3, 128, kernel_size=(3, 3), stride=(1, 1), padding=(1, 1))
BatchNorm2d(128, eps=1e-05, momentum=0.1)
ReLU(inplace=True)
Conv2d(128, 128, kernel_size=(3, 3), stride=(1, 1), padding=(1, 1))
BatchNorm2d(128, eps=1e-05, momentum=0.1)
ReLU(inplace=True)
MaxPool2d(kernel_size=2, stride=2, padding=0, dilation=1)
Conv2d(128, 256, kernel_size=(3, 3), stride=(1, 1), padding=(1, 1))
BatchNorm2d(256, eps=1e-05, momentum=0.1)
ReLU(inplace=True)
Conv2d(256, 256, kernel_size=(3, 3), stride=(1, 1), padding=(1, 1))
BatchNorm2d(256, eps=1e-05, momentum=0.1)
ReLU(inplace=True)
MaxPool2d(kernel_size=2, stride=2, padding=0, dilation=1)
Conv2d(256, 512, kernel_size=(3, 3), stride=(1, 1), padding=(1, 1))
BatchNorm2d(512, eps=1e-05, momentum=0.1)
ReLU(inplace=True)
Conv2d(512, 512, kernel_size=(3, 3), stride=(1, 1), padding=(1, 1))
BatchNorm2d(512, eps=1e-05, momentum=0.1)
ReLU(inplace=True)
MaxPool2d(kernel_size=2, stride=2, padding=0, dilation=1)
Conv2d(512, 1024, kernel_size=(3, 3), stride=(1, 1))
BatchNorm2d(1024, eps=1e-05, momentum=0.1)
ReLU(inplace=True)
MaxPool2d(kernel_size=2, stride=2, padding=0, dilation=1)
Linear(in_features=1024, out_features=10, bias=True)
Softmax(dim=1)
\end{verbatim}
\endgroup

\subsection{STL-10}
For the STL-10 dataset \cite{STL102011} we used the following network architecture \hyperlink{https://github.com/aaron-xichen/pytorch-playground}{https://github.com/aaron-xichen/pytorch-playground}:
\begingroup
\fontsize{8pt}{8pt}\selectfont
\begin{verbatim}
Conv2d(3, 32, kernel_size=(3, 3), stride=(1, 1), padding=(1, 1))
BatchNorm2d(32, eps=1e-05, momentum=0.1)
ReLU(inplace=True)
MaxPool2d(kernel_size=2, stride=2, padding=0, dilation=1)
Conv2d(32, 64, kernel_size=(3, 3), stride=(1, 1), padding=(1, 1))
BatchNorm2d(64, eps=1e-05, momentum=0.1)
ReLU(inplace=True)
MaxPool2d(kernel_size=2, stride=2, padding=0, dilation=1)
Conv2d(64, 128, kernel_size=(3, 3), stride=(1, 1), padding=(1, 1))
BatchNorm2d(128, eps=1e-05, momentum=0.1)
ReLU(inplace=True)
MaxPool2d(kernel_size=2, stride=2, padding=0, dilation=1)
Conv2d(128, 128, kernel_size=(3, 3), stride=(1, 1), padding=(1, 1))
BatchNorm2d(128, eps=1e-05, momentum=0.1)
ReLU(inplace=True)
MaxPool2d(kernel_size=2, stride=2, padding=0, dilation=1)
Conv2d(128, 256, kernel_size=(3, 3), stride=(1, 1))
BatchNorm2d(256, eps=1e-05, momentum=0.1)
ReLU(inplace=True)
Conv2d(256, 256, kernel_size=(3, 3), stride=(1, 1))
BatchNorm2d(256, eps=1e-05, momentum=0.1)
ReLU(inplace=True)
MaxPool2d(kernel_size=2, stride=2, padding=0, dilation=1)
Linear(in_features=256, out_features=10, bias=True)
Softmax(dim=1)
\end{verbatim}
\endgroup

\subsection{SVHN}
For the SVHN dataset \cite{SVHN2011} we used the following network architecture \hyperlink{https://github.com/aaron-xichen/pytorch-playground}{https://github.com/aaron-xichen/pytorch-playground}:
\begingroup
\fontsize{8pt}{8pt}\selectfont
\begin{verbatim}
Conv2d(3, 32, kernel_size=(3, 3), stride=(1, 1), padding=(1, 1))
BatchNorm2d(32, eps=1e-05, momentum=0.1)
ReLU(inplace=True)
Dropout(p=0.3, inplace=False)
Conv2d(32, 32, kernel_size=(3, 3), stride=(1, 1), padding=(1, 1))
BatchNorm2d(32, eps=1e-05, momentum=0.1)
ReLU(inplace=True)
Dropout(p=0.3, inplace=False)
MaxPool2d(kernel_size=2, stride=2, padding=0, dilation=1)
Conv2d(32, 64, kernel_size=(3, 3), stride=(1, 1), padding=(1, 1))
BatchNorm2d(64, eps=1e-05, momentum=0.1)
ReLU(inplace=True)
Dropout(p=0.3, inplace=False)
Conv2d(64, 64, kernel_size=(3, 3), stride=(1, 1), padding=(1, 1))
BatchNorm2d(64, eps=1e-05, momentum=0.1)
ReLU(inplace=True)
Dropout(p=0.3, inplace=False)
MaxPool2d(kernel_size=2, stride=2, padding=0, dilation=1)
Conv2d(64, 128, kernel_size=(3, 3), stride=(1, 1), padding=(1, 1))
BatchNorm2d(128, eps=1e-05, momentum=0.1)
ReLU(inplace=True)
Dropout(p=0.3, inplace=False)
Conv2d(128, 128, kernel_size=(3, 3), stride=(1, 1), padding=(1, 1))
BatchNorm2d(128, eps=1e-05, momentum=0.1)
ReLU(inplace=True)
Dropout(p=0.3, inplace=False)
MaxPool2d(kernel_size=2, stride=2, padding=0, dilation=1)
Conv2d(128, 256, kernel_size=(3, 3), stride=(1, 1))
BatchNorm2d(256, eps=1e-05, momentum=0.1)
ReLU(inplace=True)
Dropout(p=0.3, inplace=False)
MaxPool2d(kernel_size=2, stride=2, padding=0, dilation=1)
Linear(in_features=256, out_features=10, bias=True)
Softmax(dim=1)
\end{verbatim}
\endgroup

\subsection{USPS}\label{uspslisting}
For the USPS dataset \cite{uspsdataset} we used the following network architecture \hyperlink{https://github.com/aaron-xichen/pytorch-playground}{https://github.com/aaron-xichen/pytorch-playground}:
\begingroup
\fontsize{8pt}{8pt}\selectfont
\begin{verbatim}
Conv2d(1, 32, kernel_size=(3, 3), stride=(1, 1), padding=(1, 1))
BatchNorm2d(32, eps=1e-05, momentum=0.1, affine=True, track_running_stats=True)
ReLU(inplace=True)
MaxPool2d(kernel_size=2, stride=2, padding=0, dilation=1)
Conv2d(32, 64, kernel_size=(3, 3), stride=(1, 1))
BatchNorm2d(64, eps=1e-05, momentum=0.1, affine=True, track_running_stats=True)
ReLU(inplace=True)
MaxPool2d(kernel_size=2, stride=2, padding=0, dilation=1)
Linear(in_features=576, out_features=144, bias=True)
Dropout(p=0.25, inplace=False)
Linear(in_features=144, out_features=10, bias=True)
Softmax(dim=1)
\end{verbatim}
\endgroup

\end{document}

%% file: Include/layers/init.tex
\usetikzlibrary{quotes,arrows.meta}
\usetikzlibrary{positioning}

%% file: main.bbl
\begin{thebibliography}{10}

\bibitem{Shadish2001}
W.R. Shadish, T.D. Cook, and D.T. Campbell.
\newblock {\em Experimental and Quasi-Experimental Designs for Generalized
  Causal Inference}.
\newblock Boston: Houghton Mifflin, 2001.

\bibitem{Marwala2015}
Tshilidzi Marwala.
\newblock {\em Causality, Correlation and Artificial Intelligence for Rational
  Decision Making}.
\newblock World Scientific, Singapore, 2015.

\bibitem{Zaremba2014}
Anna Zaremba and Tomaso Aste.
\newblock Measures of causality in complex datasets with application to
  financial data.
\newblock {\em Entropy}, 16(4):2309–2349, Apr 2014.

\bibitem{Lizier2010}
Joseph~T. Lizier and Mikhail Prokopenko.
\newblock Differentiating information transfer and causal effect.
\newblock {\em The European Physical Journal B}, 73:605--615, 2010.

\bibitem{Schreiber2000}
Thomas Schreiber.
\newblock Measuring information transfer.
\newblock {\em Phys. Rev. Lett.}, 85:461--464, Jul 2000.

\bibitem{Barnett2009}
Lionel Barnett, Adam~B. Barrett, and Anil~K. Seth.
\newblock Granger causality and transfer entropy are equivalent for gaussian
  variables.
\newblock {\em Phys. Rev. Lett.}, 103:238701, Dec 2009.

\bibitem{Hlavackova-Schindler2011}
Kate{\v{r}}ina Hlav{\'a}{\v{c}}kov{\'a}-Schindler.
\newblock Equivalence of granger causality and transfer entropy: A
  generalization.
\newblock {\em Appl. Math. Sci.}, 5(73):3637--3648, 2011.

\bibitem{Massey1990}
James~L. Massey.
\newblock Causality, feedback and directed information.
\newblock In {\em Proc. 1990 Int. Symp. on Info. Th. and its Applications,
  Hawaii, USA, Nov. 27-30}, pages 303--305, 1990.

\bibitem{Cataron2017}
Angel Cațaron and Razvan Andonie.
\newblock Transfer information energy: {A} quantitative causality indicator
  between time series.
\newblock In {\em Artificial Neural Networks and Machine Learning - {ICANN}
  2017 - 26th International Conference on Artificial Neural Networks, Alghero,
  Italy, September 11-14, 2017, Proceedings, Part {II}}, pages 512--519, 2017.

\bibitem{Cataron2019}
Angel Caţaron and Răzvan Andonie.
\newblock Transfer information energy: A quantitative indicator of information
  transfer between time series.
\newblock {\em Entropy}, 20(5), 2018.

\bibitem{Lizier2011}
Joseph~T. Lizier, Jakob Heinzle, Annette Horstmann, John-Dylan Haynes, and
  Mikhail Prokopenko.
\newblock Multivariate information-theoretic measures reveal directed
  information structure and task relevant changes in fmri connectivity.
\newblock {\em Journal of Computational Neuroscience}, 30(1):85--107, Feb 2011.

\bibitem{Vicente2011}
Raul Vicente, Michael Wibral, Michael Lindner, and Gordon Pipa.
\newblock Transfer entropy---a model-free measure of effective connectivity for
  the neurosciences.
\newblock {\em Journal of Computational Neuroscience}, 30(1):45--67, Feb 2011.

\bibitem{Shimono2015}
Masanori Shimono and John~M. Beggs.
\newblock Functional clusters, hubs, and communities in the cortical
  microconnectome.
\newblock {\em Cerebral cortex (New York, N.Y. : 1991)}, 25(10):3743--3757, Oct
  2015.
\newblock 25336598[pmid].

\bibitem{HuiFang2018}
Hui Fang, Victoria Wang, and Motonori Yamaguchi.
\newblock Dissecting deep learning networks—visualizing mutual information.
\newblock {\em Entropy}, 20(11), 2018.

\bibitem{Obst2010}
Oliver Obst, Joschka Boedecker, and Minoru Asada.
\newblock Improving recurrent neural network performance using transfer
  entropy.
\newblock In {\em Proceedings of the 17th International Conference on Neural
  Information Processing: Models and Applications - Volume Part II}, ICONIP'10,
  pages 193--200, Berlin, Heidelberg, 2010. Springer-Verlag.

\bibitem{Feraud2002}
Raphael Féraud and Fabrice Clérot.
\newblock A methodology to explain neural network classification.
\newblock {\em Neural Networks}, 15(2):237 -- 246, 2002.

\bibitem{Herzog2017}
Sebastian Herzog, Christian Tetzlaff, and Florentin W{\"{o}}rg{\"{o}}tter.
\newblock Transfer entropy-based feedback improves performance in artificial
  neural networks.
\newblock {\em CoRR}, abs/1706.04265, 2017.

\bibitem{Herzog2020}
Sebastian Herzog, Christian Tetzlaff, and Florentin Wörgötter.
\newblock Evolving artificial neural networks with feedback.
\newblock {\em Neural Networks}, 123:153 -- 162, 2020.

\bibitem{Patterson2017}
Josh Patterson and Adam Gibson.
\newblock {\em Deep Learning: A Practitioner's Approach}.
\newblock O'Reilly Media, Inc., 1st edition, 2017.

\bibitem{Moldovan2020}
Adrian Moldovan, Angel Ca{\c{t}}aron, and R{\u{a}}zvan Andonie.
\newblock Learning in feedforward neural networks accelerated by transfer
  entropy.
\newblock {\em Entropy}, 22(1):102, 2020.

\bibitem{Bossomaier2016}
Terry Bossomaier, Lionel Barnett, Michael Harr\'e, and Joseph~T. Lizier.
\newblock {\em An Introduction to Transfer Entropy. Information Flow in Complex
  Systems}.
\newblock Springer, 2016.

\bibitem{Baghli2006}
Mustapha Baghli.
\newblock A model-free characterization of causality.
\newblock {\em Economics Letters}, 91(3):380 -- 388, 2006.

\bibitem{Hlavackova-Schindler2007}
Kate{\v{r}}ina Hlav{\'a}{\v{c}}kov{\'a}-Schindler, Milan Paluš, Martin
  Vejmelka, and Joydeep Bhattacharya.
\newblock Causality detection based on information-theoretic approaches in time
  series analysis.
\newblock {\em Physics Reports}, 441(1):1 -- 46, 2007.

\bibitem{Kaiser2002}
A.~Kaiser and T.~Schreiber.
\newblock Information transfer in continuous processes.
\newblock {\em Physica D: Nonlinear Phenomena}, 166(1):43 -- 62, 2002.

\bibitem{Gencaga2015}
Deniz Gencaga, Kevin~H. Knuth, and William~B. Rossow.
\newblock A recipe for the estimation of information flow in a dynamical
  system.
\newblock {\em Entropy}, 17(1):438--470, 2015.

\bibitem{Hlavackova-Schindler2009}
Kate{\v{r}}ina Hlav{\'a}{\v{c}}kov{\'a}-Schindler.
\newblock Causality in time series: Its detection and quantification by means
  of information theory.
\newblock In Frank Emmert-Streib and Matthias Dehmer, editors, {\em Information
  Theory and Statistical Learning}, pages 183--207. Springer US, Boston, MA,
  2009.

\bibitem{Zhu2015}
Jie Zhu, Jean-Jacques Bellanger, Huazhong Shu, and R\'{e}gine
  Le~Bouquin~Jeann\`{e}s.
\newblock Contribution to transfer entropy estimation via the
  k-nearest-neighbors approach.
\newblock {\em Entropy}, 17(6):4173--4201, 2015.

\bibitem{726791}
Y.~{Lecun}, L.~{Bottou}, Y.~{Bengio}, and P.~{Haffner}.
\newblock Gradient-based learning applied to document recognition.
\newblock {\em Proceedings of the IEEE}, 86(11):2278--2324, 1998.

\bibitem{Alex2012}
Alex Krizhevsky, Ilya Sutskever, and Geoffrey~E. Hinton.
\newblock Imagenet classification with deep convolutional neural networks.
\newblock In {\em Advances in Neural Information Processing Systems}, 2012.

\bibitem{Simonyan15}
Karen Simonyan and Andrew Zisserman.
\newblock Very deep convolutional networks for large-scale image recognition.
\newblock In {\em International Conference on Learning Representations}, 2015.

\bibitem{DBLP:journals/corr/SzegedyLJSRAEVR14}
Christian Szegedy, Wei Liu, Yangqing Jia, Pierre Sermanet, Scott~E. Reed,
  Dragomir Anguelov, Dumitru Erhan, Vincent Vanhoucke, and Andrew Rabinovich.
\newblock Going deeper with convolutions.
\newblock {\em CoRR}, abs/1409.4842, 2014.

\bibitem{DBLP:journals/corr/HeZRS15}
Kaiming He, Xiangyu Zhang, Shaoqing Ren, and Jian Sun.
\newblock Deep residual learning for image recognition.
\newblock {\em CoRR}, abs/1512.03385, 2015.

\bibitem{DBLP:journals/corr/abs-1905-11946}
Mingxing Tan and Quoc~V. Le.
\newblock {EfficientNet: R}ethinking model scaling for convolutional neural
  networks.
\newblock {\em CoRR}, abs/1905.11946, 2019.

\bibitem{Musat2020}
Bogdan Mu{\c{s}}at and R{\u{a}}zvan Andonie.
\newblock Semiotic aggregation in deep learning.
\newblock {\em Entropy}, 22(12):1365, 2020.

\bibitem{Rumelhart1986}
D.~E. Rumelhart, G.~E. Hinton, and R.~J. Williams.
\newblock {\em Parallel Distributed Processing: Explorations in the
  Microstructure of Cognition, Vol. 1}.
\newblock MIT Press, Cambridge, MA, USA, 1986.

\bibitem{Shalev2011}
Shai Shalev-Shwartz, Yoram Singer, Nathan Srebro, and Andrew Cotter.
\newblock Pegasos: primal estimated sub-gradient solver for svm.
\newblock {\em Mathematical Programming}, 127:3--30, 2011.

\bibitem{CIFAR10Kriz2009}
Alex Krizhevsky.
\newblock Learning multiple layers of features from tiny images.
\newblock Technical report, 2009.

\bibitem{fashionMNIST2017}
Han Xiao, Kashif Rasul, and Roland Vollgraf.
\newblock Fashion-mnist: a novel image dataset for benchmarking machine
  learning algorithms.
\newblock 2017.

\bibitem{STL102011}
Adam Coates, Andrew Ng, and Honglak Lee.
\newblock An analysis of single-layer networks in unsupervised feature
  learning.
\newblock {\em Journal of Machine Learning Research - Proceedings Track},
  15:215--223, 01 2011.

\bibitem{SVHN2011}
Yuval Netzer, Tao Wang, Adam Coates, Alessandro Bissacco, Bo~Wu, and Andrew Ng.
\newblock Reading digits in natural images with unsupervised feature learning.
\newblock {\em NIPS}, 01 2011.

\bibitem{uspsdataset}
J.~J. {Hull}.
\newblock A database for handwritten text recognition research.
\newblock {\em IEEE Transactions on Pattern Analysis and Machine Intelligence},
  16(5):550--554, 1994.

\bibitem{Lecun99}
Yann Lecun, Patrick Haffner, Léon Bottou, and Yoshua Bengio.
\newblock Object recognition with gradient-based learning.
\newblock In {\em Contour and Grouping in Computer Vision}. Springer, 1999.

\bibitem{Gilbert2013}
Charles~D Gilbert and Wu~Li.
\newblock Top-down influences on visual processing.
\newblock {\em Nature Reviews Neuroscience}, 14(5):350--363, 2013.

\bibitem{Spillmann2015}
Lothar Spillmann, Birgitta Dresp-Langley, and Chia-Huei Tseng.
\newblock Beyond the classical receptive field: the effect of contextual
  stimuli.
\newblock {\em Journal of Vision}, 15(9):7--7, 2015.

\bibitem{Ke2019}
Nan~Rosemary Ke, Olexa Bilaniuk, Anirudh Goyal, Stefan Bauer, Hugo Larochelle,
  Chris Pal, and Yoshua Bengio.
\newblock Learning neural causal models from unknown interventions, 2019.

\end{thebibliography}
